# Scope of Research on Particle Swarm Optimization Based Data Clustering

Ms.Vishakha A. Metre [1], Mr. Pramod B. Deshmukh [2]
Department of Computer Engineering
D. Y. Patil College of Engineering, Akurdi
Pune - India

## ABSTRACT

Optimization is nothing but a mathematical technique which finds maxima or minima of any function of concern in some realistic region. Different optimization techniques are proposed which are competing for the best solution. Particle Swarm Optimization (PSO) is a new, advanced, and most powerful optimization methodology that performs empirically well on several optimization problems. It is the extensively used Swarm Intelligence (SI) inspired optimization algorithm used for finding the global optimal solution in a multifaceted search region. Data clustering is one of the challenging real world applications that invite the eminent research works in variety of fields. Applicability of different PSO variants to data clustering is studied in the literature, and the analyzed research work shows that, PSO variants give poor results for multidimensional data. This paper describes the different challenges associated with multidimensional data clustering and scope of research on optimizing the clustering problems using PSO. We also propose a strategy to use hybrid PSO variant for clustering multidimensional numerical, text and image data.

## I. INTRODUCTION

The world is becoming more and more complex and competitive day by day, hence optimal decision must be taken in almost every field. Therefore optimization can be defined as an act of finding the best/optimal solution under given set of rules. Several researchers have generated different solutions for solving linear as well as non-liner optimization problems. Mathematical definition of an optimization problem consists of fitness function that describes the problem under given circumstances representing the solution space of the problem. The traditional optimization techniques use first derivatives to locate the optima, but for the ease of evaluation; various derivatives free optimization techniques have been proposed in recent years [28].

Different kinds of optimization techniques have been constructed in order to solve different optimization problems as there is no optimization technique that can be used to solve all kinds of optimization problems. The latest optimization techniques such as ant colony optimization, artificial immune systems, fuzzy optimization, genetic algorithms, neural networks, and particle swarm optimization algorithm are able to solve complex engineering problems [28][29].Swarm Intelligence (SI), being a subfield of Artificial Intelligence (AI) has proved its remarkable stand in the field of optimization. SI adapts the behavior of swarms that can be seen in fish schools, flock of birds, and in insects like midges and mosquitoes. These animal groups, mainly fish schools and bird flocks visibly show structural ordering behavior of the organisms, hence even if they change their direction and shape, they seems to travel as a single coherent entity. SI is based on five main principles of collective behavior such as homogeneity, locality, collision avoidance, velocity matching, and flock centering [27], which are depicted in Fig. 1. Different swarm intelligence inspired algorithms have been proposed to study optimization problems like NP-hard problems (e.g. clustering, data mining, Graph problems, job scheduling, network routing, Quadratic Assignment Problem, and Traveling Salesman Problem). PSO is well known optimization algorithm in the SI area till date [27].

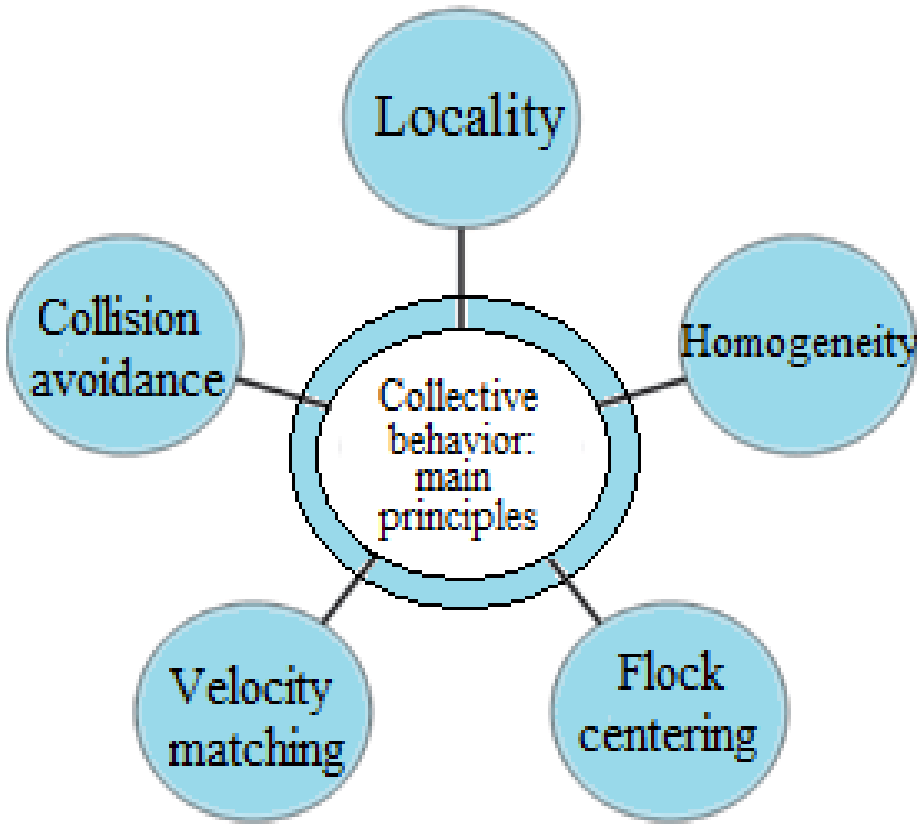

Fig. 1. The main principles of collective behavior [23]

The rest of the paper is organized as follows. In Section II, a brief literature survey on different variant of PSO is described. Section III provides the scope of research, presenting the challenges associated with clustering, PSO, and PSO based clustering. Section IV describes the studied research work, followed by Section V which gives brief overview of the data sets to be used. Finally Section VI concludes this paper.

## II. LITERATURE SURVEY

Different types of PSO variants are developed by the researchers in the field of SI. Some are advanced versions of PSO while few are hybrid versions. Many of PSO variants are modified for removing the limitations associated with it. These variants, possessing excellent features find its applications to clustering in variety of ways. Some of the promising PSO based algorithms are described in the Table I.

## III. SCOPE OF RESEARCH

In recent years, it has been recognized that the partitional clustering technique is well suited for clustering large datasets due to its relatively low computational requirements [26]. The linear time complexity of the partitioning technique makes it extensively used technique. The most familiar partitioning clustering algorithm is the K-Means algorithm and its variants. Although K-Means and other partitional clustering methods have presented interesting abilities, it still suffers from several limitations [25]. During last few years, the problem of Clustering has been approached from different disciplines. Many optimization algorithms have been developed in recent years for solving limitations of clustering. Swarm intelligence inspired algorithms are most promising amongst them. Clustering with swarm-based algorithms (PSO) is emerging as an alternative to more conventional clustering techniques. Data clustering with PSO algorithms have recently been shown to produce good results in a wide variety of real-world data [28],[29]. Following subsections are meant to illustrate the exact challenging scenario of clustering, PSO, and application of PSO to clustering; which have motivated us to acquire this research work,

### A. *k-Means/Partitional Clustering*

Data clustering is said to be as an unsupervised classification method, possessing a basic aim of dividing the given data set into the clusters having a constraint of similarities and dissimilarities. The basic assumptions of data clustering are characteristics of data, information in the data explaining the data set and its relationship with each other. The main criteria to check the accuracy of clustering is maximum inter cluster distance, minimum intra cluster distance, and minimum Root Mean Square Error (RMSE). K-Means is the oldest and most commonly used partitional clustering technique that is used in almost all application areas, although it presents following major limitations [5],[6],[20],[23],[24],[25],

- Difficult to predict ‘K’ value in advance before starting the clustering process (i.e. difficult to predict initial cluster centroids/centers and initial no. of clusters in advance).
- With global cluster, it does not work well.
- Different initial partitions may result in different final clustering solution.
- Problem of convergence to local optima.
- Dead unit problem.
- Does not work well with clusters of different size and different density (in original data).
- Selection of distance measuring function plays a very important role in proving the effectiveness and quality of clustering result.
- It is very sensitive to noise or outliers.

### B. *Particle Swarm Optimization*

PSO is a well known evolutionary computation technique that simulates the choreography of birds within a flock in the search of globally optimized solution [30]. It follows a stochastic optimization method inspired by SI. The basic idea of PSO is the one in which each and every particle in the swarm represents an individual solution while entire swarm represents the solution space. The PSO starts with searching in parallel, having a group of particles and approaches to the optimum solution with its current velocity, its previous best velocity, and the previous best velocity of its neighbors. The PSO algorithms adapt the strategy of birds searching for food in the space. As PSO is simple, effective, fast, and derivative free, it is the most suitable algorithm till date for solving complex optimization problems, but it has some limitations which are stated as below [19],[20],[22],

- It shows poor results for large search space as well as complex data sets.
- Due to high convergence speed, in terms of high dimensional search space (i.e. as the search space size increases), it prematurely converges to local optima before reaching the global optima.
- Weak local search ability (i.e. particles are flown through single point which is randomly determined by pbest & gbest positions and this point is not guaranteed to be a local optima).
- Stagnation problem due to premature convergence (i.e. if the particles are continuously converging prematurely and it is observed that there is no improvement over several time steps, such kind of phenomenon is called as stagnation).

### C. *PSO based Clustering*

Data clustering is one of the important real world application areas whereas PSO is the most promising swarm intelligence inspired optimization technique. Hence the combinations of these two domains make a challenging and most fascinating area for research work. Several PSO variants are proposed in the literature to optimize the clustering results. PSO is the great choice to improve the performance of the clustering. It is used as an evolutionary technique to evaluate the clustering process in order to get feasible solution. With the help of PSO or its variants, we observed that it can resolve the problems associated with the standard partitional clustering methods such as K-Means which are listed below [15],[16],[17],[18],[21],

- The number of clusters k (being the solution space dimension), should be specific in advance.
- As the complexity of clustering scheme increases, the method tends to trap in local optima.
- With multidimensional data sets, shows poor results with overlapping cluster relationship of particles (i.e.

TABLE I
Comparative Analysis of Different PSO based Clustering Algorithms

| **Publication** | **Paper Title** | **Algorithm/ Techniques** | **Purpose** | **Datasets/ Benchmark functions** | **PROS** | **CONS** |
|---|---|---|---|---|---|---|
| Springer 2012 S. Rana [5] | A boundary restricted adaptive PSO for data clustering | BRAPSO | To apply Boundary restriction strategy for improved clustering. | Art1, Art2, CMC, Crude oil, Glass, Iris, Vowel, and Wine. | 1. Efficient, robust and fast convergence.<br>2. Handles particles outside search space boundary. | - |
| IJORCS 2013 Mariam El-Tarabily [6] | A PSO-Based Subtractive Data Clustering Algorithm | Hybrid Subtractive clustering PSO | To achieve fast and efficient clustering. | Iris, Wine, Yeast | 1. High convergence speed and minimum fitness value.<br>2. Dimension reduction approach is not required. | 1. Overlapping cluster membership problem.<br>2. Premature convergence problem. |
| IJSER June 2013 Chetna Sethi [7] | A Linear PCA based hybrid k-Means PSO algorithm for clustering large dataset | PCA-K-PSO, k-means, Linear PCA, PSO | For clustering high dimensional data. | Artificial data sets | 1. Good clustering.<br>2. Low computational cost. | 1. Dimension reduction may degrade clustering results. |
| IEEE Transaction April 2010 Serkan Kiranyaz [8] | Fractional Particle Swarm Optimization in Multidimensional Search Space | MD-PSO with FGBF, Fuzzy clustering, FGBF, MD-PSO | To resolve premature convergence problem for multidimensional data. | Artificial data sets | 1. Appropriate for multi dimensional data sets.<br>2. Avoids early convergence to local optima. | 1. Occasional over-clustering can be encountered.<br>2. Increased no. of dimension increases cost. |
| Springer July 2010 Bara'a Ali Attea [9] | A fuzzy multi-objective particle swarm optimization for effective data clustering | FMOPSO, Multiobjective optimization Technique, Fuzzy clustering | To concentrate on the problem of multicluster membership. | Cancer, Iris, Soybean, Tic-tac-toe, Zoo | 1. Consistent and efficient clustering algorithm.<br>2. Good clustering result. | 1. Poor results with multidimensional data sets. |
| IJCSNS Jan 2008 R.Karthi [10] | Comparative evaluation of PSO Algorithm for Data Clustering using real world data sets. | GA, DE, B-PSO, C-K PSO, S-E PSO, R-PSO, C-S PSO, E-S PSO | To evaluate PSO algorithms for data clustering problems using real word data sets. | Iris, Vowel, Cancer, Glass | 1. Comparative study of the performance of different PSO variants. | 1. As the number of functional evaluation increases, no PSO variant dominates all the others on all benchmark data sets. |
| IEEE ICSP 2002 Xiao-Feng Xie [11] | Adaptive Particle Swarm Optimization on Individual Level | Adaptive PSO | To adapt swarm at individual level. | Rosenbrock, Rastrigrin, Griewank | - | 1. Adaptive method may be also used for other evolutionary computation tech., e.g. GA. |
| IEEE 2013 Jianchao Fan [12] | Cooperative Coevolution for Large-scale Opti. Based on K-Fuzzy Cluste.& Variable Trust Region Methods. | FT-DNPS, Kernel Fuzzy Clustering, Dynamic Neighbor topology, Trust Region Method. | To propose a novel methodology for large scale optimization. | Sphere, Rosenbrock's Ackley's, Griewanks's, Rastrigin, Weierstrass,Noncontinuous, Schwefel's | 1. Avoids premature convergence, Better performance. | - |
| IMECS 2012 Li-Yeh Chuang [13] | An Improved PSO for Data Clustering | Guass chaotic map based PSO | To avoid premature convergence problem. | Vowel, Iris, CMC,Cancer, Crude oil, Wine | 1. Converges to global optima with minimum error rate. | 1. Based on random selection of initial cluster centers.<br>2. Dead unit problem. |
| ICSRS Dec2009 K. Premalatha [14] | Hybrid PSO and GA for Global Maximization | Hybrid PSO + GA | To overcome convergence due to stagnation. | Shaffer, Rosenbrock, Rastrigin, Ackley. | 1. Reliable, flexible, robust method. | 1. If no change in *gbest* then stagnation occurs. |

multicluster membership problem).

- In an evaluation process of PSO for clustering, some particles have tendency to travel beyond the search space area in the search of global solution. These particles attract other particles and misguide them by spreading their pbest and gbest information which is inaccurate information since the particles are outside the boundary, which results into poor quality of clustering.

## IV. RESEARCH METHODOLOGY

We have studied and analyzed the different challenging aspects of clustering as well as PSO based clustering in previous section. We have also observed the applicability of different variants of PSO to solve clustering drawbacks. Fig. 2 hence shows the studied hybrid research procedure to solve clustering limitations. Partitional clustering methods (e.g. K-Means) or PSO based clustering methodologies either chooses the initial cluster centers and predicts the number of cluster in advance at random manner or simply ignore this condition which lead to inaccurate clustering results. Hence, if the mentioned approach is able to have preknowledge of initial cluster centers and number of clusters, the clustering process can proceed further with good initial start and ends with accurate clustering solution [1][2][25].

Amongst several clustering techniques such as K-Means, Fuzzy C-Means, Mountain clustering, and Subtractive clustering etc., we found Subtractive Clustering (SC) as the most promising one [6],[25].

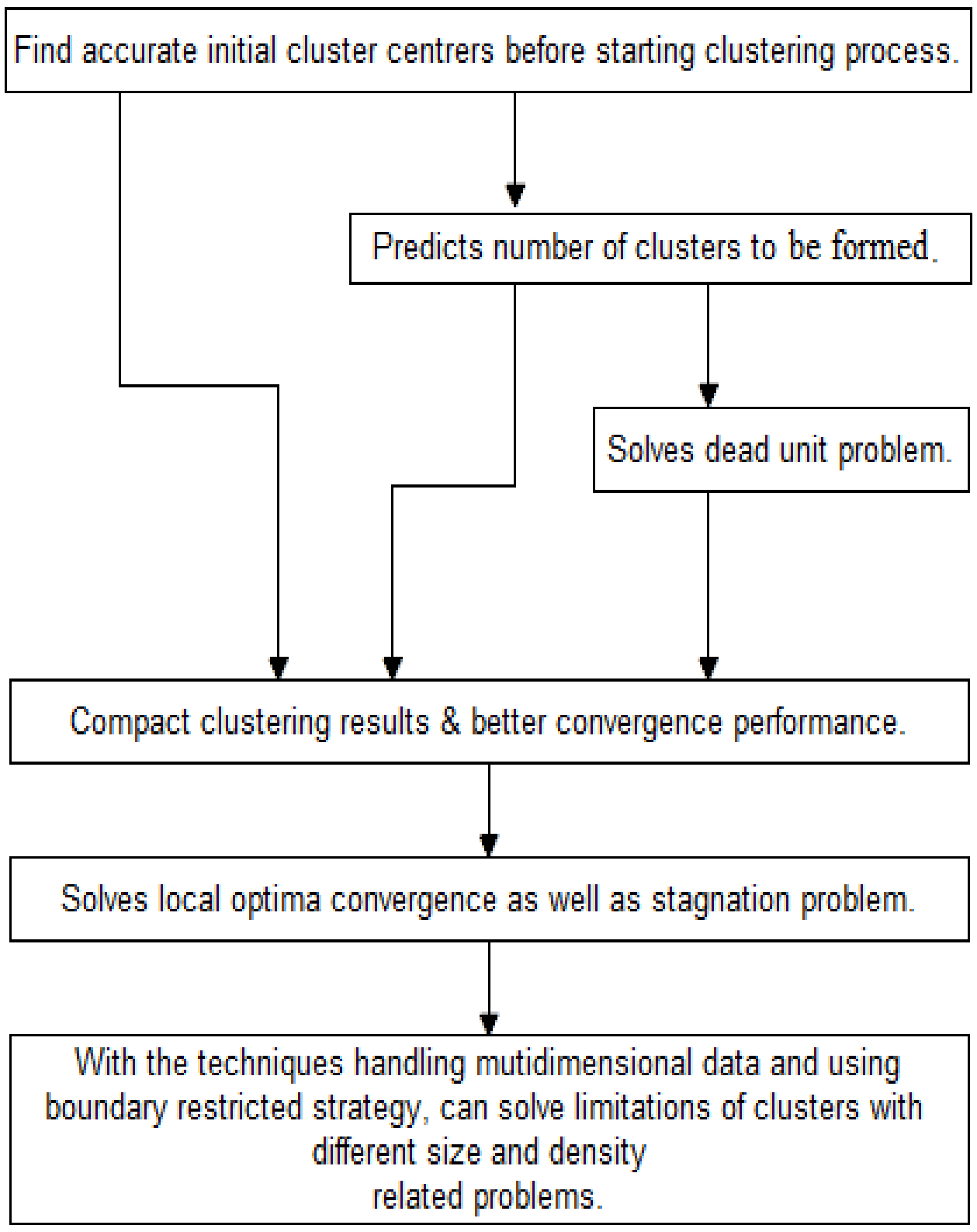

Fig. 1. Studied research procedure to solve clustering limitations [4]

### A. *Subtractive Clustering Algorithm*

*1) Input:* Collection of data points in M-dimensional space (e.g. shown in Fig. 3.)

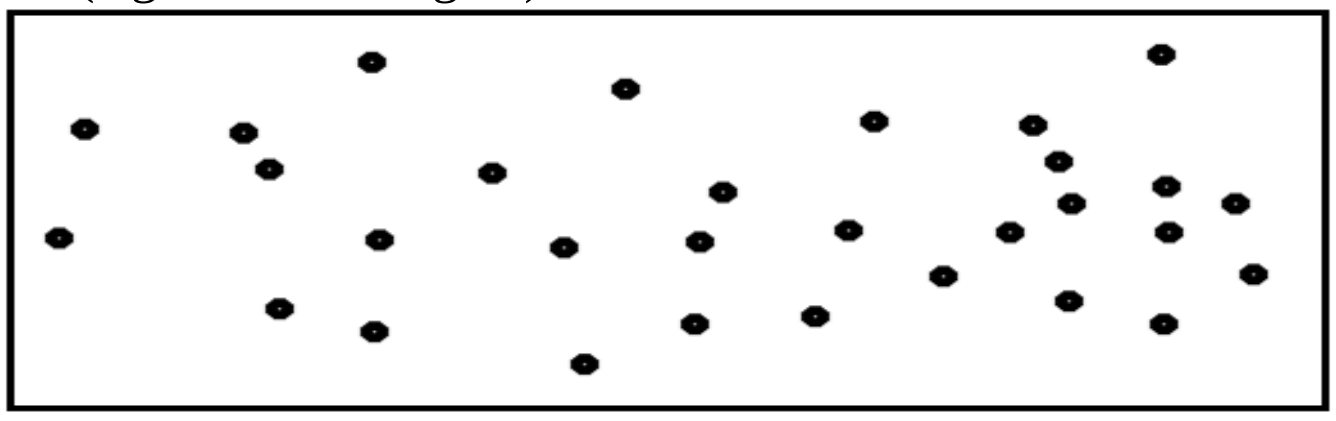

Fig. 3. Input data points[3][4]

*2) SC can be described as below:*

- It can deal with high/multidimensional data sets.
- Data points are considered as candidates for the cluster centers.
- It predicts an optimal number of clusters as well as initial cluster centers for the next phase with the help of density measurement concept [6].
- Its computational complexity is directly proportional to the number of data points in the data set and independent of the dimensions of the data set under consideration.
- It is fast and requires minimum iterations to find out cluster centers.
- It converges quickly with lower fitness value.

*3) Output:* Initial cluster centers and number of clusters (e.g. shown in Fig. 4.)

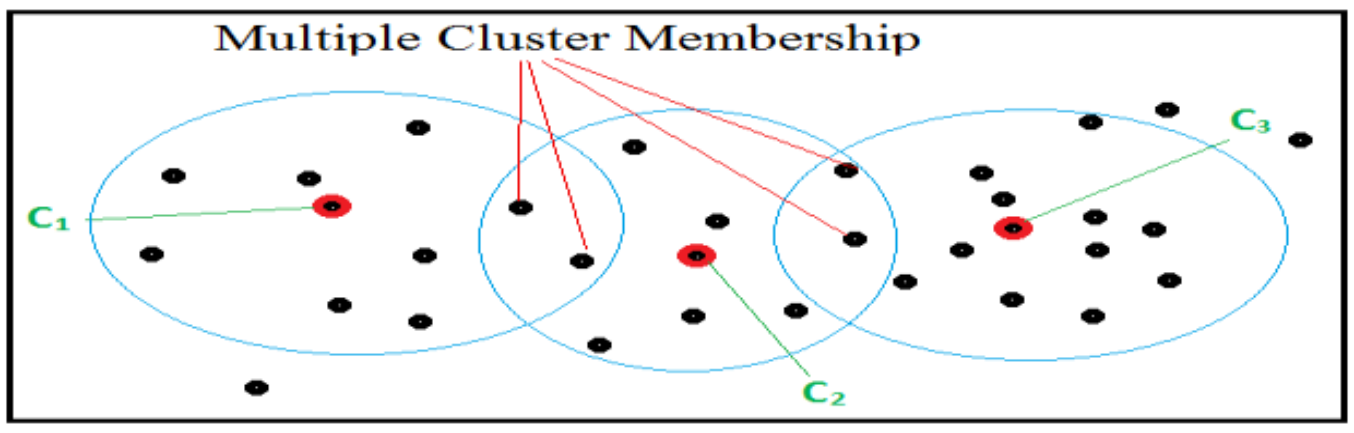

Fig. 4. Subtractive clustering output [3][4]

Similarly there are various promising variants of PSO available in the literature, but a Boundary Restricted Adaptive Particle Swarm Optimizations (BRAPSO) algorithm is the best amongst them [5].

### B. *Boundary Restricted Adaptive PSO*

*1) Input:* Initial cluster centers and number of clusters (e.g. shown in Fig. 4.)

*2) BRAPSO can be illustrated as below:*

- It uses a boundary restricted strategy [5] for those particles that go outside the search space boundary.
- This algorithm brings back the particles outside the boundary into search space area.
- It considers maximum value of the inertia weight 'w' at the beginning of the process but as the process carries out further, it decreases the inertia weight.

- It uses the concept of non-linear inertia weight adaption function because it is observed that better solution can be seen near the global solution with the small change in inertia weight factor.
- BRAPSO balances the local and global searching ability of particles in the swarm effectively.
- Fast convergence and efficient as well as effective performance is observed when compared with K-Means, K-PSO, K-NM-PSO, LDWPSO, and ALDWPSO algorithms.
- It gives minimum intra cluster distances and maximum inter cluster distances.
- It resolves multicluster membership problem of particles within swarm and avoids premature convergence problem.

***3) Final Output:*** Optimal clusters (e.g. shown in Fig. 5.)

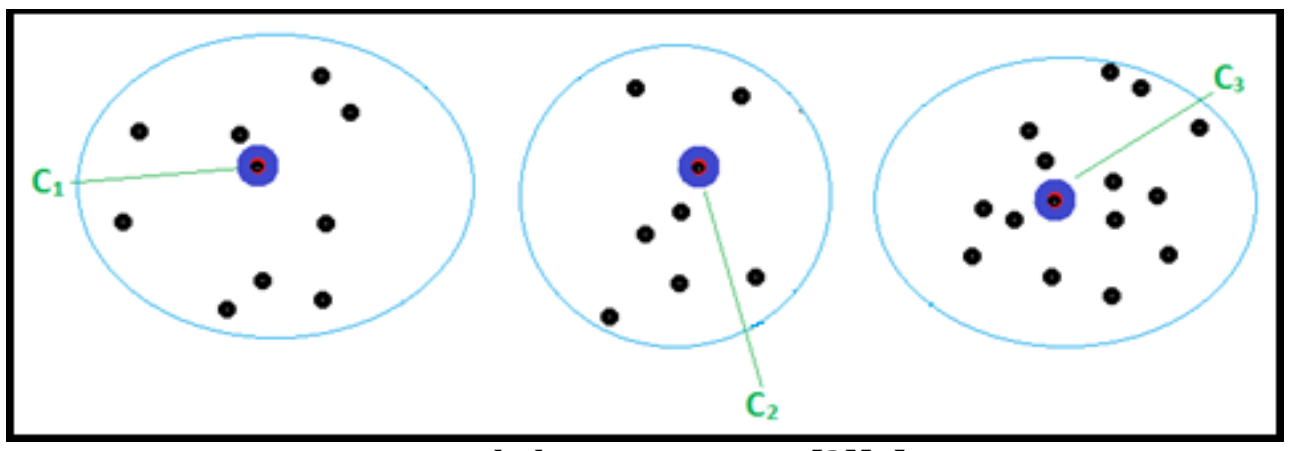

Fig. 5. Final clustering output [3][4]

Hence, hybridizing the Subtractive Clustering [2] and BRAPSO [1] algorithm will help us in optimizing the multidimensional numerical, text and image data clusters with minimum error rate. This approach will not only resolve the drawbacks of these two algorithms but also improves the clustering performance as compared to existing ones. SC-BR-APSO algorithm is presented in Fig. 6.

**SC-BR-APSO Algorithm**

1. Consider any *M*-dimensional data set having *N* particles.
2. Calculate density function,

$$D_i = \sum_{j=1}^{n} exp\left(-\frac{\|x_i - x_j\|^2}{(r_a/2)^2}\right),\ i = 1, \dots, n,\ j = 1, \dots, z$$

   /* Where $n$ and $z$ are numbers of data sets and clusters, respectively and $x_i$ and $x_j$ are $i^{th}$ and $j^{th}$ data points, respectively and $D_i$ is the density value of $i^{th}$ data point and $r_a$ is the neighborhood radius (positive constant). */
3. If sufficient numbers of clusters then identify initial cluster centers and proceed else go to step 2.
4. Randomly initialize particle velocity and position.
5. Calculate fitness function and set $p_{best}$ and $g_{best}$.
6. Calculate the value of inertia weight factor $w$ exponentially.
7. Update velocity and position of particle with boundary restriction strategy,

   *If position <= upbound and position >= lwbound*
   *Then position = position*
   *Else position = position - velocity;*

   /* Where *position* and *velocity* are the current position and velocity of particle, respectively and *upbound* and *lwbound* are the upper limit and lower limit specified for search space area. */
8. Assign particles to their nearest cluster center and recalculate the cluster centers.
9. Repeat steps until convergence.

Fig. 6. SC-BR-APSO Algorithm [3]

# V. EXPERIMENTAL STUDY

The SC-BR-APSO algorithm is intended to address and solve the limitations of clustering as well as PSO. For experimental setup, we have analyzed and studied 5 multidimensional data sets. These 5 data sets are standard datasets from UCI Repository namely CMC [5],[13], Glass [5],[10], Iris [5],[7],[9],[10],[13], Pima [17], and Wine [5],[6],[13] which are briefly described as below,

## *A. Contraceptive Method Choice (CMC)*

- Total number of instances - 1473
- Number of dimensions - 9
- Number of classes - 3
- Number of instances in each class - 629,334,510

## *B. Glass Identification Data set*

- Total number of instances - 214
- Number of dimensions - 9
- Number of classes - 6
- Number of instances in each class - 70,17,76,13,9,29

## *C. Iris Data set*

- Total number of instances -150
- Number of dimensions - 4
- Number of classes - 3
- Number of instances in each class - 50,50,50

## *D. Pima Data set*

- Total number of instances - 768
- Number of dimensions - 8
- Number of classes - 2
- Number of instances in each class - 500,268

## *E. Wine Data set*

- Total number of instances - 178
- Number of dimensions -13
- Number of classes - 3
- Number of instances in each class - 59,71 48

In order to measure and compare the performance of the hybrid algorithm used in this survey and research with the existing ones on these 5 data sets, we can consider three criteria's which are Sum of Intra Cluster Distances (SICD), Error rate, and Convergence rate. To test the performance of the hybrid algorithm on text datasets, we are working on text datasets of different dimensions namely Badges (1 attribute, 24 instances), Dresses_Attribute_Sales (13 attributes, 501 instances), Miskolc IIS Hybrid IPS (67 attributes, 1540 instances), Northex (200 attributes, 115 instances), Amazon Commerce Reviews Set (10000 attributes, 1500 instances). All these standard datasets are freely available in UCI Repository. For Badges and Dresses_Attribute_Sales datasets, SC-BR-APSO algorithm has proved its worthiness and encourages the employment of the discussed hybrid approach for image clustering.

# VI. CONCLUSION AND FUTURE WORK

We have studied and analyzed the different aspects of scope of research on the challenges associated with Data Clustering, Particle Swarm Optimization, and PSO based data clustering. We have also analyzed the different clustering as well as PSO based algorithms in order to optimize the problems associated with clustering of multidimensional data. Thus, we recommend a hybrid algorithm involving Subtractive Clustering algorithm and Boundary Restricted Adaptive PSO which is named as SC-BR-APSO algorithm for text, numerical and image data clustering. For comparison purpose, we have considered three important factors which are namely SICD, error rate, and convergence rate. SC-BR-APSO tested on 5 multidimensional numerical data sets and 2 text datasets and achieved good results. Future work predicts to achieve compact clustering result with increased convergence rate and decreased error rate as compared to existing algorithms for image datasets by means of image processing techniques.